\documentclass[11pt]{article}
\usepackage{acl}
\usepackage{times}
\usepackage{latexsym}
\usepackage[T1]{fontenc}
\usepackage[utf8]{inputenc}
\usepackage{microtype}
\usepackage{inconsolata}
\usepackage{graphicx}
\usepackage{booktabs}
\usepackage{multirow}
\usepackage{tabularx}
\usepackage{array}
\usepackage{enumitem}
\usepackage[most]{tcolorbox}
\usepackage{hyperref}
\title{Large Language Models for Causal Relations Extraction in Social Media:\\
A Validation Framework for Disaster Intelligence}
\author{
\textbf{Ujun Jeong\textsuperscript{1}},
\textbf{Saketh Vishnubhatla\textsuperscript{1}},
\textbf{Bohan Jiang\textsuperscript{1}},\\
\textbf{Andre Harrison\textsuperscript{2}},
\textbf{Adrienne Raglin\textsuperscript{2}},
\textbf{Huan Liu\textsuperscript{1}}
\\[0.5em]
\textsuperscript{1}Arizona State University, Tempe, Arizona, USA \\
\textsuperscript{2}DEVCOM Army Research Laboratory, Adelphi, Maryland, USA
\\[0.5em]
\small
\texttt{
\{ujeong1, svishnu6, bjiang14, huanliu\}@asu.edu
}
\\
\small
\texttt{
\{andre.v.harrison2.civ, adrienne.raglin2.civ\}@army.mil
}
}
\begin{document}
\maketitle
\begin{abstract}
During disasters, extracting causal relations from social media can strengthen situational awareness by identifying factors linked to casualties, physical damage, infrastructure disruption, and cascading impacts. However, disaster-related posts are often informal, fragmented, and context-dependent, and they may describe personal experiences rather than explicit causal relations. In this work, we examine whether Large Language Models (LLMs) can effectively extract causal relations from disaster-related social media posts. To this end, we (1) propose an expert-grounded evaluation framework that compares LLM-generated causal graphs with reference graphs derived from disaster-specific reports and (2) assess whether the extracted relations are supported by post-event evidence or instead reflect model priors. Our findings highlight both the potential and risks of using LLMs for causal relation extraction in disaster decision-support systems.
\end{abstract}
\section{Introduction}
Causality is a crucial yet underexplored dimension of disaster intelligence, as highlighted in a 2021 systematic review~\cite{wiegmann2020opportunities}. Agencies such as NOAA produce detailed post-event reports documenting hazards, impacts, and contributing factors. However, these reports are retrospective and labor-intensive, and they often provide only limited insight into the rapidly evolving dynamics of disasters~\cite{kryvasheyeu2016rapid}.

In contrast, social media platforms such as Twitter provide real-time disaster information through user observations, warnings, and damage reports~\cite{vieweg2010microblogging}. Hurricanes Harvey and Irma, for example, demonstrated how social media posts captured impacts such as flooding, transportation disruptions, and power outages as events unfolded, highlighting the potential of crowd-sourced data to reveal early signals of cause--effect relations during disasters~\cite{king2018social}.

However, extracting causal relations from social media remains challenging. Posts are often short, informal, fragmented, and context-dependent, with causality frequently implied rather than explicitly stated. Rule-based methods that rely on markers such as ``caused by'' or ``led to'' therefore miss many relevant relations. Although NLP methods have advanced structured event and knowledge extraction~\cite{liu2023event, lu2021text2event}, many still depend on human-annotated templates or domain-specific schemas, limiting their generalization to complex and rapidly changing disaster contexts.

Large Language Models (LLMs) offer a promising alternative because they can process noisy text, reason over incomplete context, and infer unstated relations~\cite{hao2026sere}. Yet, this capability introduces a key risk: LLMs may generate plausible causal links from prior knowledge rather than evidence in the posts. This motivates our research question: \textit{``Can LLMs extract valid causal relations from social media posts about disaster events?''}

To address this question, we develop an evaluation framework that compares LLM-generated causal graphs with expert-grounded reports and examines whether model outputs are supported by post-level evidence or influenced by prior knowledge. This framework enables three contributions:
\begin{itemize}[leftmargin=*,nosep]
\item \textbf{Expert-grounded causal graphs:} We construct disaster-specific ground-truth causal graphs by aligning an impact-chain framework with post-event reports to retain expert-grounded evidence.
\item \textbf{Social-media causal extraction:} We evaluate LLMs to extract causal relations among a fixed set of variables for social media analysis, distinguishing this from open-ended causal discovery.
\item \textbf{Evidence versus prior knowledge:} We examine how social media post quality affects LLM's causal extraction and reveal the risk of plausible but ungrounded outputs driven by model priors.
\end{itemize}
\section{Related Work}
\subsection{Social Media for Disaster Intelligence}
Social media has been widely used for disaster intelligence, including event detection, damage assessment, crisis classification, geolocation, and situational awareness~\cite{ma2025camo, vishnubhatla2025assessing, zou2023social,imran2016twitter}. CrisisMMD and HumAID provide annotated disaster posts for relevance and impact classification~\cite{alam2018crisismmd,alam2021humaid}. Prior approaches range from keyword and location-based clustering~\cite{atefeh2015survey,becker2011beyond} to neural models that capture semantic and temporal patterns~\cite{zhao2017eventfact,lee2021newevent,zhou2022weak}. \citeauthor{jiang2024grace} leveraged GPT-2 to analyze social media posts but without validation.

\subsection{Causal Relation Extraction from Text}
Causal relation extraction has been extensively studied, from lexical cue-based systems to neural event extraction frameworks~\cite{liu2023event,lou2023universal,lu2021text2event,susanti2025paths}. Recent work has also explored using LLMs for causal graph construction from natural text. Closest to our setting, \citeauthor{saklad2025can} evaluate whether LLMs can infer causal graphs from academic documents, requiring external judgment to align the generated graph with the reference graph.

Our work is distinct in two ways. First, we use an impact-chain framework to define the target variables and their level of abstraction in advance, which removes ambiguity about node equivalence and enables direct graph evaluation. Second, to the best of our knowledge, this is the first study to extract disaster-related causal relationships from social media and validate them using post-event evidence documented in standardized expert reports.
\section{Building Ground Truth Causal Graph}
\subsection{Expert Evidence Matching and Record}
Ground-truth causal graphs for specific disaster events, defined here as sets of directed \textit{(Cause, Effect)} relations, are not readily available. To address this gap, we adopt a \textit{match-and-record} approach that combines established disaster-management frameworks with event-specific expert evidence~\cite{zebisch2021vulnerability}. We start with a general causal graph based on the impact-chain framework for each disaster type, then refine it using causality evidence from post-event reports.

\begin{enumerate}
    \item \textbf{Initial variables and edges:}
We build on the impact-chain framework, which represents causal pathways among hazards, exposure, vulnerability, and impacts in disaster risk analysis~\cite{zebisch2021vulnerability,zebisch2022climate}. Specifically, we use the storm-related framework proposed by \citeauthor{pittore2023border} as the initial set of variables and edges to ensure the level of abstraction.

    \item \textbf{Collect post-event expert evidence:}
For each disaster event, we consult official, standardized expert reports, such as NOAA's post-event reports~\cite{cangialosi2021irma,cangialosi2018harvey}, rather than aggregating ad hoc sources. These reports provide expert-grounded evidence of causal relations specific to the disaster event.

    \item \textbf{Evidence-based pruning:}
We retain only the edges that are explicitly supported by textual evidence in the reports and remove unsupported relations. The causality evidence is managed in a tabulated document for review.
\end{enumerate}

\begin{figure}[t]
    \centering
    \includegraphics[width=0.5\textwidth]{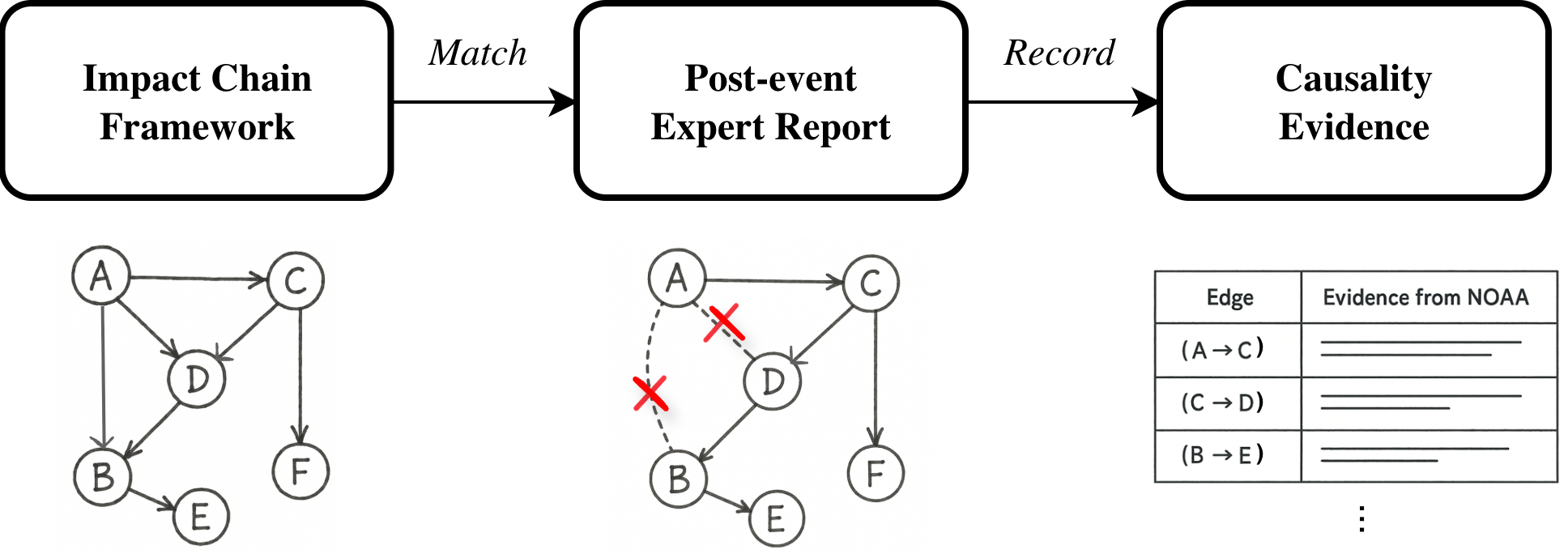}
\caption{Matching impact-chain with disaster-specific post-event report to record ground-truth causal relations.}
\label{fig:causal_ground_truth}
\end{figure}

Figure~\ref{fig:causal_ground_truth} summarizes the construction of the ground-truth causal graphs from the impact-chain framework and authoritative post-event reports. The resulting graphs and supporting evidence tables were reviewed by two Ph.D.-level researchers in disaster response and prediction, and will be released with the data sources upon acceptance.
\subsection{Disaster Cases and Data Curation}
We focus on disasters that satisfy two criteria: (1) the availability of publicly accessible social media data and (2) the existence of standardized, authoritative post-event reports that enable consistent causal analysis. Using the CrisisMMD~\cite{alam2018crisismmd} and HumAID~\cite{alam2021humaid} benchmarks, we identified Hurricanes Irma and Harvey as the only events meeting both requirements. This constraint led our study to focus on U.S.-based disasters documented by NOAA reports, which provide structured, event-specific evidence of causal mechanisms. After removing duplicate post IDs, the curated datasets contain 9,824 posts for Hurricane Irma and 10,662 posts for Hurricane Harvey.
\section{Causal Relation Extraction}
\subsection{LLMs and Causal Graph Generation}
We compare LLMs along two dimensions: (1) access to social media data and (2) model openness, distinguishing between closed-source and open-weight systems. This design allows us to examine whether platform-specific contextual knowledge and model scale affect causal relation extraction.
\begin{itemize}
    \item \textbf{\texttt{Grok4.3}}: Closed-source LLM by xAI with reasoning and native social media access for historical and real-time disaster-related posts.
   
    \item \textbf{\texttt{GPT-5.5}}: Closed-source LLM by OpenAI with reasoning and long-context capabilities, but no native social-media access.
   
    \item \textbf{\texttt{Mistral-7B}}: Open-weight LLM by Mistral AI representing smaller, resource-constrained models with limited context windows.
\end{itemize}
Because smaller models such as Mistral-7B have limited context windows, we process posts in batches of 20 rather than as a single input. This approach follows a chunking-based strategy for long-context LLM processing, where smaller segments are analyzed independently and later combined. For each batch, the model extracts directed causal pairs, which are then aggregated across batches with duplicate edges merged~\cite{ratner2023parallel}. In practice, batching also helps mitigate the lost-in-the-middle effect that can occur when processing long-context prompts~\cite{friedman2022unstructured}.
\subsection{Prompt Template Standardization}
The prompt design follows established practices in instruction-based information extraction by specifying the task, constraining the output schema, and grounding the extraction in event-specific context. In this study, the scope of causal relations is restricted to the target variables defined in the impact-chain framework for each disaster event type. Each model receives three inputs: the disaster event type, the list of canonical target variables from the impact-chain framework, and a batch of posts. The prompt instructs the model to identify causal mentions in the posts, map them to the supplied variables, and output directed causal pairs.
\begin{tcolorbox}[
  colback=black!5!white,
  colframe=black!75!black,
  title={Prompt Template: Causal Graph Generation},
  fonttitle=\small\bfseries,
  sharp corners=south,
  enhanced,
  boxrule=0.8pt,
  coltitle=white
]
\small
\label{prompt:causal_graph_generation}
\textbf{Task:} Identify cause and effect relations from social media posts related to \textit{[Disaster Event]}.

\textbf{Instructions:}
\begin{itemize}
\setlength{\itemsep}{0.0em}
\item If available, conduct a native social media search for posts related to \textit{[Disaster Event]}; otherwise, rely solely on the provided posts.
\item Restrict all causes and effects to these variables: \textit{[Variables for Disaster Type]}.
\item Extract causal relations that are explicitly stated or reasonably implied in the posts.
\item Represent each causal relation as a directed edge in the format of \textit{(Cause, Effect)}.
\end{itemize}
\vspace{0.25em}
\textbf{Input:} \textit{[A batch of social media posts]}.
\vspace{0.25em}

\textbf{Output:} A list of causal relations.
\end{tcolorbox}
\subsection{Evaluation Metrics}
To align with standard practices in causal discovery, we evaluate the generated causal graphs using the comparison metrics adopted by \citeauthor{saklad2025can}. These include node and edge precision and recall, F1 score, structural Hamming distance (SHD), and normalized SHD (nSHD). Due to space constraints, detailed definitions are provided in Appendix~\ref{metric:causaldiscovery}.

When interpreting these metrics, it is important to note that our setting differs from that of \citeauthor{saklad2025can}. In their setup, LLMs generate both variables and edges from long-form text, requiring an external judge (e.g., another LLM) to resolve semantic mismatches between the generated and reference variables. In contrast, our evaluation provides a fixed set of canonical variables based on the impact chain framework. Our task is limited to identifying the relevant variables to include and extracting whether causal relations exist between them.
\begin{table*}[t]
\centering
\renewcommand{\arraystretch}{1.05}
\setlength{\tabcolsep}{4pt}
\tiny
\begin{tabular*}{\textwidth}{@{\extracolsep{\fill}} c c c c c c c c c}
\toprule
\textbf{Disaster Event} (Type) & \textbf{Model}
& \textbf{Node Precision ($\uparrow$)}
& \textbf{Node Recall ($\uparrow$)}
& \textbf{Edge Precision ($\uparrow$)}
& \textbf{Edge Recall ($\uparrow$)}
& \textbf{F1 ($\uparrow$)}
& \textbf{SHD ($\downarrow$)}
& \textbf{nSHD ($\downarrow$)} \\
\midrule
\multirow{4}{*}{\shortstack{\textbf{Hurricane Irma} \\ (Tropical Cyclone)}}
    & \texttt{Random} & 0.34$\pm$0.24 & 0.30$\pm$0.29 & 0.06$\pm$0.10 & 0.08$\pm$0.15 & 0.12$\pm$0.10 & 31.2$\pm$22.10 & 0.56$\pm$0.39 \\
    & \texttt{Mistral-7B} & 0.67$\pm$0.11 & 0.88$\pm$0.08 & 0.21$\pm$0.09 & 0.20$\pm$0.09 & 0.46$\pm$0.08 & 17.2$\pm$2.40 & 0.31$\pm$0.04 \\
    & \texttt{GPT-5.5} & 0.85$\pm$0.04 & 1.00$\pm$0.00 & 0.64$\pm$0.06 & 0.45$\pm$0.07 & 0.71$\pm$0.03 & 9.60$\pm$1.02 & 0.17$\pm$0.02 \\
    & \texttt{Grok4.3} & \textbf{0.93$\pm$0.05} & 1.00$\pm$0.00 & 0.63$\pm$0.08 & \textbf{0.57$\pm$0.03} & \textbf{0.75$\pm$0.03} & 9.40$\pm$1.50 & 0.17$\pm$0.03 \\
\midrule
\multirow{4}{*}{\shortstack{\textbf{Hurricane Harvey} \\ (Tropical Cyclone)}}
    & \texttt{Random} & 0.34$\pm$0.24 & 0.30$\pm$0.29 & 0.05$\pm$0.10 & 0.09$\pm$0.18 & 0.12$\pm$0.10 & 30.5$\pm$21.77 & 0.54$\pm$0.39 \\
    & \texttt{Mistral-7B} & 0.84$\pm$0.05 & 0.78$\pm$0.09 & 0.36$\pm$0.14 & 0.20$\pm$0.09 & 0.52$\pm$0.07 & 12.6$\pm$1.50 & 0.23$\pm$0.03 \\
    & \texttt{GPT-5.5} & 0.94$\pm$0.04 & 0.83$\pm$0.06 & 0.66$\pm$0.13 & 0.37$\pm$0.03 & 0.66$\pm$0.04 & 9.20$\pm$1.40 & 0.16$\pm$0.03 \\
    & \texttt{Grok4.3} & \textbf{0.98$\pm$0.05} & \textbf{0.94$\pm$0.06} & 0.62$\pm$0.06 & \textbf{0.44$\pm$0.09} & \textbf{0.71$\pm$0.05} & 9.10$\pm$1.04 & 0.16$\pm$0.02 \\
\bottomrule
\end{tabular*}
\caption{LLM performance on causal relation extraction from disaster-related posts across 10 runs. Boldface denotes significantly better in performance between \texttt{Grok4.3} and \texttt{GPT-5.5} based on paired t-tests ($p < 0.05$).}
\label{tab:disaster-models-compact}
\vspace{1em}
\begin{tabular*}{\textwidth}{@{\extracolsep{\fill}} c c c c c c c c c}
\toprule
\textbf{Disaster Event} (Type) & \textbf{Model}
& \textbf{Node Precision ($\uparrow$)}
& \textbf{Node Recall ($\uparrow$)}
& \textbf{Edge Precision ($\uparrow$)}
& \textbf{Edge Recall ($\uparrow$)}
& \textbf{F1 ($\uparrow$)}
& \textbf{SHD ($\downarrow$)}
& \textbf{nSHD ($\downarrow$)} \\
\midrule
\multirow{3}{*}{\shortstack{\textbf{Hurricane Irma} \\ (Tropical Cyclone)}}
    & \texttt{Mistral-7B} & 0.94$\pm$0.08 & 0.50$\pm$0.24 & 0.17$\pm$0.15 & 0.05$\pm$0.13 & 0.34$\pm$0.13 & 12.4$\pm$0.80 & 0.22$\pm$0.01 \\
    & \texttt{GPT-5.5} & 1.00$\pm$0.00 & 0.60$\pm$0.12 & 1.00$\pm$0.00 & 0.28$\pm$0.04 & 0.58$\pm$0.07 & 8.60$\pm$0.49 & 0.15$\pm$0.01 \\
    & \texttt{Grok4.3} & 1.00$\pm$0.00 & 0.60$\pm$0.09 & 0.80$\pm$0.18 & 0.25$\pm$0.04 & 0.54$\pm$0.04 & 9.90$\pm$0.94 & 0.18$\pm$0.02 \\
\midrule
\multirow{3}{*}{\shortstack{\textbf{Hurricane Harvey} \\ (Tropical Cyclone)}}
    & \texttt{Mistral-7B} & 0.83$\pm$0.14 & 0.43$\pm$0.13 & 0.43$\pm$0.23 & 0.11$\pm$0.07 & 0.35$\pm$0.11 & 11.0$\pm$0.63 & 0.20$\pm$0.01 \\
    & \texttt{GPT-5.5} & 1.00$\pm$0.00 & 0.50$\pm$0.11 & 0.47$\pm$0.07 & 0.13$\pm$0.04 & 0.41$\pm$0.07 & 11.2$\pm$0.40 & 0.20$\pm$0.01 \\
    & \texttt{Grok4.3} & N/A & N/A & N/A & N/A & N/A & N/A & N/A \\
\bottomrule
\end{tabular*}
\caption{Controlled evaluation of LLMs' prior-knowledge reliance using non-informative posts across 10 runs. N/A indicates that the model refuses to generate a causal graph due to insufficient evidence in the non-informative posts.}
\label{tab:disaster-models-ablation}
\end{table*}
\section{Results and Analysis}
\subsection{Extraction from Relevant Disaster Posts}
To confirm that LLMs extract meaningful causal relations from social media posts rather than relying on random guessing, we compared their performance against a random baseline. This random model generates causal graphs using the Erdős--Rényi model~\cite{erd6s1960evolution} over a fixed set of canonical variables provided in the prompt, connecting pairs with a probability of 0.5. As shown in Table~\ref{tab:disaster-models-compact}, all evaluated LLMs substantially outperform the random baseline across both disaster events. These findings demonstrate that the models benefit from understanding social media posts to infer disaster-specific causal relations.

Among LLMs, \texttt{Grok4.3} shows the highest F1 scores (0.75 on Irma and 0.71 on Harvey), driven mainly by higher edge recall while maintaining strong node precision. This superior performance arises from the combination of both its advanced reasoning capabilities and native access to social media data. \texttt{GPT-5.5} performs competitively and leads among models without native social media access, though its lower edge recall reflects a more conservative extraction approach. \texttt{Mistral-7B} shows the weakest overall performance, with notably lower edge precision and recall. Differences across models appear primarily at the edge level rather than in node identification, as the prompt already supplies the canonical variable set.

\subsection{LLMs' Reliance on Prior Knowledge}
While LLMs can generate coherent causal graphs, their reasoning may rely more on internal priors than on the provided data. To evaluate this effect, we conducted a controlled experiment using only non-informative posts from the CrisisMMD dataset for Hurricane Irma ($  n=677  $) and Hurricane Harvey ($  n=799  $). Since these posts contain no relevant information to disasters and any external search functions were disabled, any inferred causality reflect the models’ knowledge priors.

As shown in Table~\ref{tab:disaster-models-ablation}, model performance drops substantially in the ablation setting, particularly in edge recall and overall F1, indicating that LLMs fall back on pre-trained priors when social media evidence is insufficient. \texttt{GPT-5.5} maintains relatively higher F1 scores (0.58 on Irma and 0.41 on Harvey), suggesting greater reliance on prior knowledge to produce plausible structures. \texttt{Mistral-7B} exhibits the weakest performance overall. In contrast, \texttt{Grok4.3} refuses to generate any causal graph for Harvey and displays high selectivity on Irma, reflecting stricter evidence thresholds and better calibration against unsupported inferences. This behavior highlights the need of conservative grounding mechanism before high-stakes disaster intelligence deployment.

\section{Conclusion and Future Work}
We proposed a framework to validate LLM-generated causal relations extracted from social media posts against a ground-truth graph built from expert disaster reports. Results show that \texttt{Grok4.3} achieves the highest F1 score using native social media access. While \texttt{GPT-5.5} performs closely among models relying on provided disaster posts, our ablation study highlights a key limitation: without sufficient grounding evidence, even advanced models rely on internal priors to generate plausible but unverified causal links. Therefore, despite the promise LLMs show for disaster-specific extraction, robust evidence-grounding is essential before their deployment in decision-support systems.

We will extend the framework to other disaster types as corresponding NOAA reports become available, investigate the temporal causal relations on social media as crises unfold~\cite{vishnubhatla2025interventional, vishnubhatla2025assessing, sheth2021causebox}, and develop a human-AI collaborative verification pipeline to enhance the reliability of extracted causal graphs.

\section*{Limitations}
Our study is limited to disasters with standardized NOAA reports, which constrains the evaluation primarily to U.S.-based events. Although other disasters, such as earthquakes in CrisisMMD and HumAID, are well documented, their reports are often distributed across heterogeneous sources (e.g., EERI and regional agencies), making the consistent construction of expert-grounded causal graphs challenging. In addition, models differ not only in capability but also in data-access conditions. Accordingly, our evaluation reflects each model together with its accessible data sources, rather than isolating intrinsic reasoning ability alone. In particular, xAI’s access restrictions prevent downloading certain social-media inputs (e.g., posts retrieved by \texttt{Grok4.3}), limiting cross-model comparisons under identical social-media access conditions. To address the concern that performance differences may be driven solely by data quality, we additionally conducted a controlled experiment in which \texttt{Grok4.3} was evaluated using the same public post benchmark; the results are provided in Appendix~\ref{app:grok-public-only}.
\section*{Ethical Considerations}
Social media posts can be noisy, incomplete, and unevenly distributed across communities and locations~\cite{jeong2026user, jeong2025navigating,jeong2025fediverse, jeong2024descriptor, jeong2023user, jeong2024exploring, jeong2022nothing, jeong2022classifying, jeong2025reinforcement}. Posts may contain subjective experience, so the extracted causal graphs should be used only as decision-support tools rather than standalone decision systems. Practical deployment requires source validation, cross-referencing with official emergency communications, and expert review of high-impact relationships. For user privacy, this work uses social media posts only to construct aggregated data for disaster analysis and does not perform user profiling or infer private attributes.

\section*{Acknowledgment}
This work supported by, or in part by the U.S. Army Materiel Command under Grant Award Number W911NF24-2-0175 and by the U.S. Army Research Laboratory under Grant Award Number W911NF2020124. The views and conclusions contained in this document are those of the authors and should not be interpreted as representing the official policies of the U.S. Army Materiel Command or the U.S. Army Research Laboratory.

\bibliography{custom}
\appendix
\section{\texttt{Grok4.3} without Social-Media Access}
\label{app:grok-public-only}
To separate the effect of native social-media access from the effect of model capability, we conduct an additional diagnostic experiment. \texttt{Grok4.3} receives exactly the same public post batches used for \texttt{GPT-5.5}, bypassing native social-media access. As shown in Table~\ref{tab:grok43_metrics_two}, \texttt{Grok4.3} shows comparable performance to \texttt{GPT-5.5} in F1 scores in this controlled setting. This result indicates that the public posts in the benchmark contain sufficient causal information to support expert-aligned graph construction, yet native access is more advantageous.
\begin{table}[tph]
\centering
\small
\caption{Evaluation for \texttt{Grok4.3} across two disasters when the public-post benchmark is provided as input.}
\begin{tabular}{lcc}
\hline
\textbf{Metric} & \textbf{Hurricane Irma} & \textbf{Hurricane Harvey} \\
\hline
\textbf{Node Precision} & 0.89$\pm$0.00 & 0.91$\pm$0.11 \\
\textbf{Node Recall} & 0.99$\pm$0.04 & 0.83$\pm$0.06 \\
\textbf{Edge Precision} & 0.62$\pm$0.07 & 0.71$\pm$0.08 \\
\textbf{Edge Recall} & 0.43$\pm$0.05 & 0.42$\pm$0.04 \\
\textbf{F1} & 0.72$\pm$0.03 & 0.68$\pm$0.02 \\
\textbf{SHD} & 10.0$\pm$1.00 & 8.40$\pm$0.49 \\
\textbf{nSHD} & 0.18$\pm$0.02 & 0.15$\pm$0.01 \\
\hline
\end{tabular}
\label{tab:grok43_metrics_two}
\end{table}
\section{Causal Relation Extraction Metrics}
\label{metric:causaldiscovery}
We evaluate causal graph construction at the node and edge levels. Let $G_{\mathrm{ref}}=(V_{\mathrm{ref}},E_{\mathrm{ref}})$ denote the expert-grounded reference graph and $G_{\mathrm{pred}}=(V_{\mathrm{pred}},E_{\mathrm{pred}})$ denote the LLM-generated graph, where nodes represent causal variables and directed edges represent causal relations. For graph elements $X\in\{V,E\}$, precision and recall are defined as follows:
\[
P_X=\frac{|X_{\mathrm{ref}}\cap X_{\mathrm{pred}}|}{|X_{\mathrm{pred}}|},
\qquad
R_X=\frac{|X_{\mathrm{ref}}\cap X_{\mathrm{pred}}|}{|X_{\mathrm{ref}}|}.
\]
Node-level scores are obtained with $X=V$, and edge-level scores with $X=E$. Precision measures the validity of generated elements, while recall measures coverage of the reference causal graph.

We compute an overall micro-averaged F1 score across both nodes and edges as
\[
F_1=\frac{2(|V_{\mathrm{ref}}\cap V_{\mathrm{pred}}| + |E_{\mathrm{ref}}\cap E_{\mathrm{pred}}|)}{|V_{\mathrm{ref}}| + |E_{\mathrm{ref}}| + |V_{\mathrm{pred}}| + |E_{\mathrm{pred}}|}.
\]
We also report structural Hamming distance (SHD), defined as the number of edge edits required to transform $G_{\mathrm{pred}}$ into $G_{\mathrm{ref}}$. Let
\[
\mathcal{R}=\{(u,v)\in E_{\mathrm{ref}} : (v,u)\in E_{\mathrm{pred}}\}
\]
denote the set of reversed causal relations. Then,
\[
\mathrm{SHD}
=
|E_{\mathrm{pred}}\setminus E_{\mathrm{ref}}|
+
|E_{\mathrm{ref}}\setminus E_{\mathrm{pred}}|
-
|\mathcal{R}|.
\]
We normalize SHD by the number of possible directed edges among reference variables:
\[
\mathrm{nSHD}
=
\frac{\mathrm{SHD}}{|V_{\mathrm{ref}}|(|V_{\mathrm{ref}}|-1)}.
\]
Subtracting $|\mathcal{R}|$ ensures that each reversed edge is counted as one edit rather than separately as a false positive and a false negative.
\end{document}